%% file: A_MSC2016.tex
\documentclass[letterpaper, 10 pt, conference]{ieeeconf}  

\IEEEoverridecommandlockouts                              

\usepackage{graphics} 
\usepackage{graphicx}
\usepackage{epsfig} 
\usepackage{times} 
\usepackage{amsmath} 
\usepackage{amssymb}  
 \usepackage{algorithm}
\PassOptionsToPackage{noend}{algpseudocode}
\usepackage{algpseudocode}
\algnewcommand\And{\textbf{ and }}
\algnewcommand\Or{\textbf{ or }}

\usepackage[font=small]{caption}
\usepackage{subcaption}

\usepackage{fontenc}
\usepackage[utf8]{inputenc}
\usepackage{mathtools}
\usepackage{color}
\usepackage{citesort}

\usepackage{epstopdf}
\usepackage{tabularx}
\usepackage{hhline}
\usepackage{url}

\DeclareMathAlphabet{\pazocal}{OMS}{zplm}{m}{n}

\newcommand{\Vs}{\pazocal{V}}

\title{\LARGE \bf
Technical Report: Optimal Surveillance of Dynamic Parades using Teams of Aerial Robots
}

\author{Kostas Alexis
\thanks{$^{1}$K. Alexis is with with the University of Nevada, Reno,
	1664 N. Virginia Street, Reno, NV 89557, USA
	{\tt\small kalexis@unr.edu}}%
}

\begin{document}

\maketitle
\thispagestyle{empty}
\pagestyle{empty}

\begin{abstract}
This technical report addresses the problem of optimal surveillance of the route followed by a dynamic parade using a team of aerial robots. The dynamic parade is considered to take place within an urban environment, it is discretized and at every iteration, the algorithm computes the best possible placing of the aerial robotic team members, subject to their camera model and the occlusions arising from the environment. As the parade route is only as well covered as its least--covered point, the optimization objective is to place the aerial robots such that they maximize the minimum coverage over the points in the route at every time instant of it. A set of simulation studies is used to demonstrate the operation and performance characteristics of the approach, while computational analysis is also provided and verifies the good scalability properties of the contributed algorithm regarding the size of the aerial robotics team. 
\end{abstract}

\section{INTRODUCTION}

Aerial robotics have demonstrated their ability to provide rapid coverage of complex areas and environments by exploiting miniaturized sensing technology and their advanced locomotion capabilities. Nowadays, aerial robots of very limited cost present robust flight behavior~\cite{alexis2015robust,APST_ECC_14}, and can be equipped with a multi--modal sensing suite that may contain visible light cameras~\cite{JANOSCH_PAPER_INSP,nikolic2014synchronized,PAT_JINT_2015}, thermal imaging~\cite{Oettershagen_FSR2015,OMMRLSLKS_ICRA_15,Doherty_VictimDetection,SEARCH_AND_RESCUE_2008} or even Light Detection and Ranging (LiDAR) devices~\cite{zhang2014loam_loam} and more. At the same time, progress in robotic perception has enabled the online, real--time, $3\textrm{D}$ reconstruction of the environment~\cite{omari2014visual,lynen2013robust,omari2015dense}, tracking of areas and targets of interest~\cite{PAT_ISVC_2015} or even semantic scene understanding~\cite{fernandez2015uav}. Finally, the sucessful combination of modern path planning strategies with the real--time localization and mapping capabilities of the robot has allowed aerial robots to navigate or even explore autonomously in possibly cluttered, challenging and previously unknown environments~\cite{SIP_AURO_2015,bircher_robotica,BABOOMS_ICRA_15,NBVP_ICRA_16,bircher2016receding,APST_MSC_2015,papachristos2016augmented,papachristos2016distributed}. 

\begin{figure}[h!]
\centering
  \includegraphics[width=0.99\columnwidth]{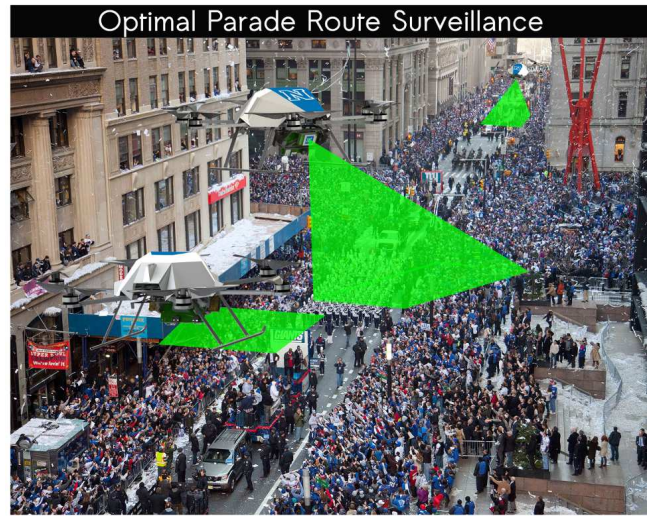}
\caption{Motivation figure of the optimal multi--aerial robot parade route surveillance algorithm: a team of aerial robots could be requested to provide coverage of a complex social event such as a parade in New York. }
\label{fig:motivation}
\end{figure}

Aiming to further leverage these outstanding achievements, this work deals with the challenge of using aerial robots to monitor dynamic social phenomena such as parades taking place in our cities. In particular, we aim to address the problem of optimally coordinating and positioning a team of aerial robots --each of them equipped with a camera sensor-- such that they can provide optimal surveillance of a dynamically evolving parade route taking place within an urban environment. The parade route is able to change its spatial distribution and form dynamically, the aerial robots are subject ot the limitations of their sensing modules and the goal is to optimize the totally achieved coverage along the parade route. As the parade route is only as well covered as its least--covered point, the optimization objective is to place the aerial robots of the team such that they maximize the minimum coverage over the points in the route at every time instant of it. Figure~\ref{fig:motivation} presents the motivation behind the algorithmic contribution of our work.

%
%

To approach this problem, we contribute an algorithm that considers a team of aerial robots capable of flying holonomic trajectories and equipped with a camera sensor of limited field of view, assumes a dynamically evolving parade route within an urban environment consisting of buildings or other occlusion structures and aims to find the best possible ``guarding'' positions of the robot team such that optimal coverage is provided at every instance of the parade. As the parade route evolves, the robot team modifies its position to provide the best coverage at any time. As this problem is in general nonconvex and NP--hard, we contribute an algorithm that provides approximate solutions via convexification very fast. To demonstrate the capabilities of the algorithm we present a set of simulation studies, while the computational properties of the algorithm are also analyzed. 

The rest of this document is organized as follows: Section~\ref{sec:problem} overviews and details the specific problem considered, while Section~\ref{sec:algorithm} describes the proposed optimal multi--aerial robot dynamic parade route surveillance algorithm. Subsequently, Section~\ref{sec:sim} presents detailed simulation results and computational analysis of the algorithm. Finally, conclusions are drawin Section~\ref{sec:concl}.

\section{PROBLEM DESCRIPTION}\label{sec:problem}

A dynamic parade following the route trajectory $\mathbf{r}(t)$ is considered to take place in an urban $2\textrm{D}$ map, subsets of which are occupied by buildings--obstacles of rectangle shape. Given a set of $S$ aerial robots capable of flying holonomic trajectories and a sensor model constrained by a field--of--view $\textrm{FOV}$, the problem is to find the set of aerial robot trajectories $\mathbf{T}(t) = [\mathbf{t}_1(t),\mathbf{t}_2(t),...,\mathbf{t}_N(t)]$ that \emph{maximize the minimum coverage} for every point of $\mathbf{r}(t)$. Due to its nature as a problem of finding dynamic guard positions to ensure coverage of a desired subset of the environment, this problem is expected to be NP--hard. The inclusion of visibility constraints and obstacles in the environment is then further complicating the effort to derive optimal solutions.

\section{PROPOSED ALGORITHM}\label{sec:algorithm}

The specifics of the algorithmic approach to solve the problem of optimal dynamic parade route surveillance using a team of aerial robots are provided below.

\subsection{Representation of the Parade Route}

The dynamic parade is considered to follow the route $\mathbf{r}(t)$ and its dynamic evolution is sampled every $T_s$. The time--trajectory $\mathbf{r}(t)$ is considered to be constructed via the sequential, piecewise connection of linear segments, an approach that allows to easily model a parade that takes place within an urban environment. For every $k$--th sample of the route $\mathbf{r}(k)$, it is discretized into a set of $m$ points $\{r^1(k),r^2(k),...,r^m(k)\}$.

\subsection{Optimization Objective}

For the parade route $\mathbf{r}(t)$, its discretized version $\mathbf{r}(k)$ per $k$--th iteration of the algorithm, and considering a team of  $S$ aerial robots, an equal amount of ``guard positions'' is desired to be computed to optimize the parade coverage. These guard positions may be selected from an arbitrary large set of possible guard locations $n$ (fixed or varying per iteration). For these guard positions, the decision variable $x_g(k) \in \{0,1\}^n$ is defined and becomes $x_g(k)^i = 1$ if and only if a robot is placed at the $i$--th location. Associated with each robot location $i$ is a coverage vector $\alpha_i(k)\in \mathbf{R}^m$, which describes how well an aerial robot placed at location $i$ would cover each point in the current of the current instance $\mathbf{r}(k)$ of the parade route. Assuming additive coverage, the vector that describes the total coverage of every edge will be defined by $\mathbf{A}(k)\mathbf{x}_g(k)$, where $\mathbf{A}(k)\in\mathbf{R}^{m\times n}$ has $\alpha_i(k)$ as its $i$--th column. 

Subsequently, as the parade route is only as secure as its least well--covered point, the optimization problem that deals with how to optimally position the aerial robots for the $k$--th sample of the route $\mathbf{r}(k)$ takes the form:

\begin{eqnarray}
 &&\max t \\ \nonumber
 \textrm{s.t.~} && t\le \mathbf{A}(k)\mathbf{x}_g(k),\\ \nonumber && \mathbf{x}_g(k)\in\{0,1\}^n, \\ \nonumber && \mathbf{1}^T\mathbf{x}_g(k)=S
\end{eqnarray}

\noindent This problem is nonconvex and, in general, NP--hard due to the necessary boolean decision variable in its definition. This fact necessitates the derivation of methods and approaches that can approximate the optimal solutions efficiently, while presenting superior performance characteristics. This can be achieved via appropriate relaxations leading to the convexification of the problem~\cite{Loefberg_ARCP,cvxbook}. 

\subsection{Relaxation for Convexification}\label{sec:relaxation}

In order to perform an efficient --yet accurate-- convexification of the problem, we form the following convex \emph{relaxation}:

\begin{eqnarray}
 &&\max t \\ \nonumber
 \textrm{s.t.~} && t\le \mathbf{A}(k)\mathbf{x}_g(k),\\ \nonumber && 0\le \mathbf{x}_g(k)\le 1,\\ \nonumber && \mathbf{1}^T\mathbf{x}_g(k)=S
\end{eqnarray}
\noindent by constraining $\mathbf{x}_g(k)\in [0,1]^n$. In general, the solution to this relaxed problem, $\mathbf{x}_g^\star$ will have fractional variables~\cite{cvxbook}. As a boolean allocation is considered in order to specifically assign a guard location to every robot, the \emph{iterated weighted} $\ell_1$ \emph{heuristic} will be used to achieve the recovery of a Boolean solution~\cite{cvxbook}.

\subsection{Application of the Iterated Weighted $\ell_1$ Heuristic}	

In order to recover a boolean solution, an approach is to solve a sequence of convex problems where the linear term $-\mathbf{w}^T\mathbf{x}_g(k)$ is added to the objective, and then picking the weight vector $\mathbf{w}\in\mathbf{R}^n_+$ at each iteration to try and induce a sparse solution vector $\mathbf{x}_g^\star (k)$. Enhancing sparsity via reweighted $\ell_1$ optimization is an extensively employed approach in convex optimization. Broadly, given a set $\mathbf{v}$ and denoting its \emph{cardinality} as $card(\mathbf{v})$, the iterated weighted $\ell_1$ heuristic is the process of minimizing $card(\mathbf{v})$ over $\mathbf{v}\in \Vs$ through the following process: 

\begin{algorithm}[h]
\label{alg:loneheur}
\begin{algorithmic}[1]
\State $\omega = 0$
\While{running}
	\State{minimize $||\mathbf{diag}(\omega)\mathbf{v}||_1$ over $\mathbf{v}\in \Vs$}
	\State{$\omega_i = 1/(\epsilon + \|v_i|)$}
\EndWhile
\end{algorithmic}
\end{algorithm}

Naturally, this process is extended for the case of matrices, while the matrix rank operator $\mathbf{rank}(\cdot)$ is then acting with the role of the cardinality operator. For the problem of finding solution to the relaxed, convex, problem of Section~\ref{sec:relaxation}, the iterated $\ell_1$ heuristic consists of initializing $\mathbf{w} = 0$ and repeating the two steps:

\noindent \emph{Step 1:}
\begin{eqnarray}
  &&\max t-\mathbf{w}^T \mathbf{x}_g(k) \\ \nonumber 
  \textrm{s.t.~} && t\le \mathbf{A}(k)\mathbf{x}_g(k),\\ \nonumber && 0\le\mathbf{x}_g(k)\le 1, \\ \nonumber && \mathbf{1}^T\mathbf{x}_g(k)=S
\end{eqnarray}

\noindent \emph{Step 2:}
\begin{eqnarray}
 \textrm{Let~} w_i = \alpha/(\tau + x_g^i),\forall i
\end{eqnarray}

\noindent Until a Boolean solution is reached. Within these expressions, $\alpha$ and $\tau$ are adjusted to promote a sparse solution. Typical choices would be $\alpha =1$ and $\tau=10^-4$.  Intuitively, the weight vector $\mathbf{w}$ is incentivizing elements of $\mathbf{x}_g(k)$ which were close to zero in the last iteration towards zero in the next iteration. It is highlighted that the $\ell_1$ heuristic is characterized by increased performance as it typically converges within $5$ or fewer iterations. 

\subsection{Iterative Algorithm Execution}

The aforementioned steps provide the solution of placing a team of aerial robots at the optimal guard positions to ensure the best coverage of a fixed instane of the parade route. As the parade is in fact dynamic, these steps are executed iteratively. At every step $k$ --sampled at a possibly varying sampling period $T_s$-- the current instance of the route $\mathbf{r}(k)$ is used and the relevant optimal robot positions are computed. The reference commands to the robots are then provided to the team on a nearest neighbor fashion.

\section{SIMULATION STUDIES AND ANALYSIS}\label{sec:sim}

To verify and evaluate the functionality of the algorithm, a set of simulation studies are considered. Within those, a $2\textrm{D}$ city is considered and parades are designed to follow complex trajectories within the city building blocks. At the same time, we varied the number of robots as well as the number of potential guard positions sampled in the environment. Below, a subset of these results will be presented and the computational analysis will be summarized. 

\begin{figure}[htbp]
\centering
  \includegraphics[width=0.99\columnwidth]{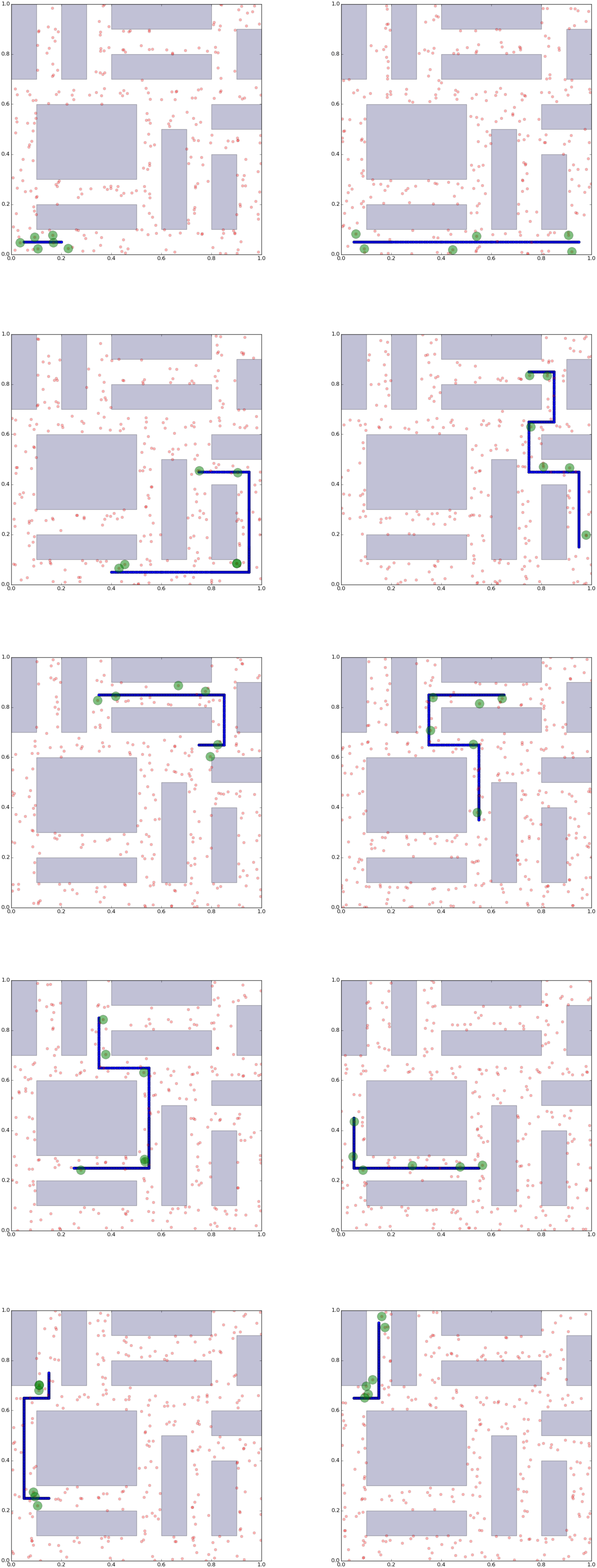}
\caption{Simulation study for the case of $6$ robots monitoring a dynamic parade route. The parade is considered to be taking place within a city--like environment consisting of $10$ building blocks. Camera-occlusions are accounted for, while the field of view of the camera that equips every robot is considered to be $175\textrm{deg}$. For this study $512$ possible guard locations are sampled within the obstacle--free subset of the world.}
\label{fig:res1}
\end{figure}

\begin{figure}[htbp]
\centering
  \includegraphics[width=0.90\columnwidth]{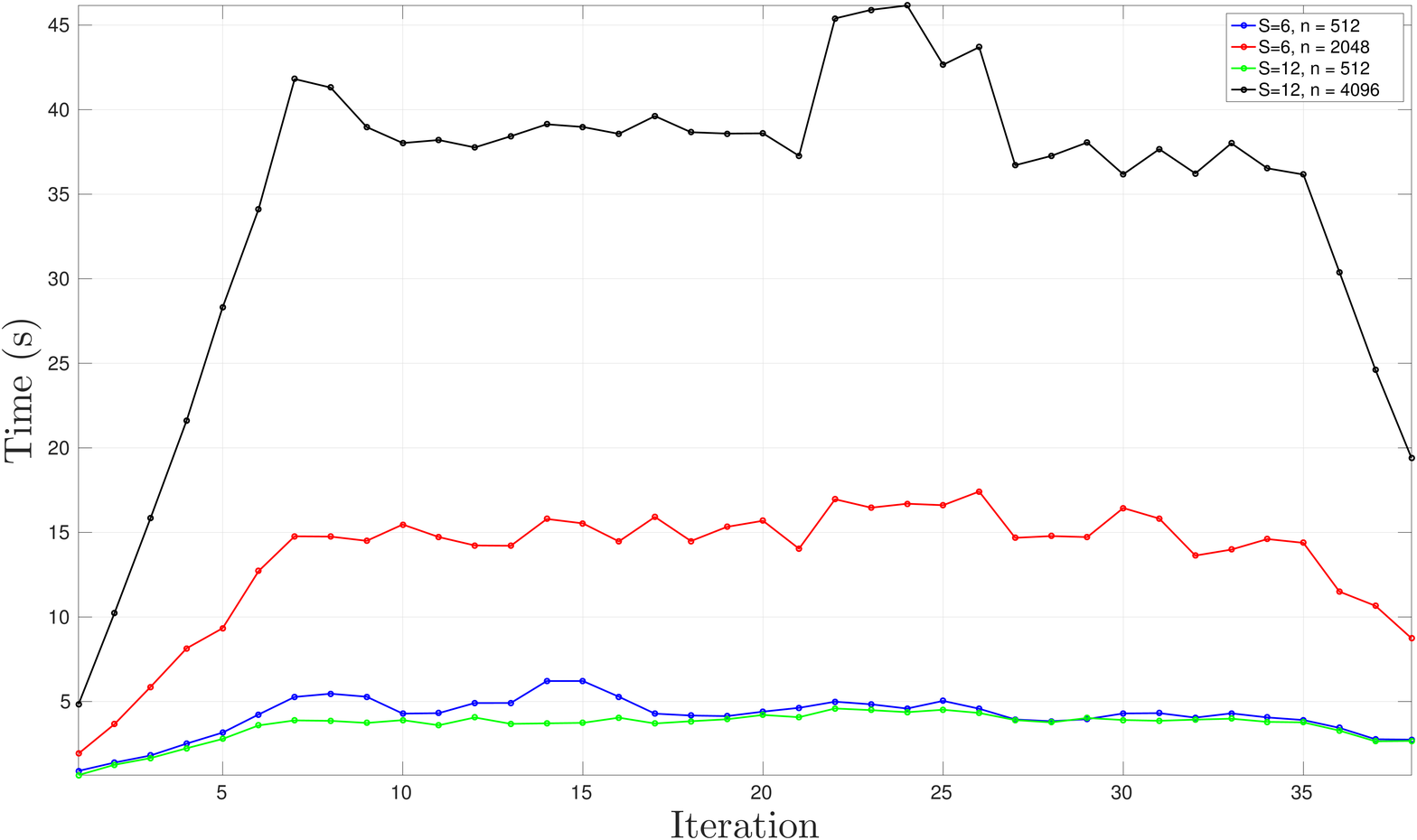}
\caption{Analysis of the computational cost per iteration of the algorithm for the aforementioned four cases utilizing $6$ or $12$ robots and different sizes of potential guard positions sets. As illustrated, the factor that greatly impacts computational time is the size of the set of possible guard locations. }
\label{fig:res_time}
\end{figure}

Figure~\ref{fig:res1} presents the case of a $6$ aerial robots commanded to monitor a complicated parade route traveling within the a city environment consisting of $10$ building blocks. The dynamic trajectory of the parade is discretized to $k_1,...,k_{37}$ samples and a total of $512$ possible guard positions are sampled within the obstacle--free subset of the workspace of the problem. Each robot is considered to be equipped with a camera with horizontal field of view $FOV=175\textrm{deg}$. As shown, the algorithm dynamically adapts the positions of the robots to find feasible, full--coverage solutions at all times. Figure~\ref{fig:res_time} presents the computation characteristics of the solution per step of iteration.

%
%

\begin{figure}[htbp]
\centering
  \includegraphics[width=0.99\columnwidth]{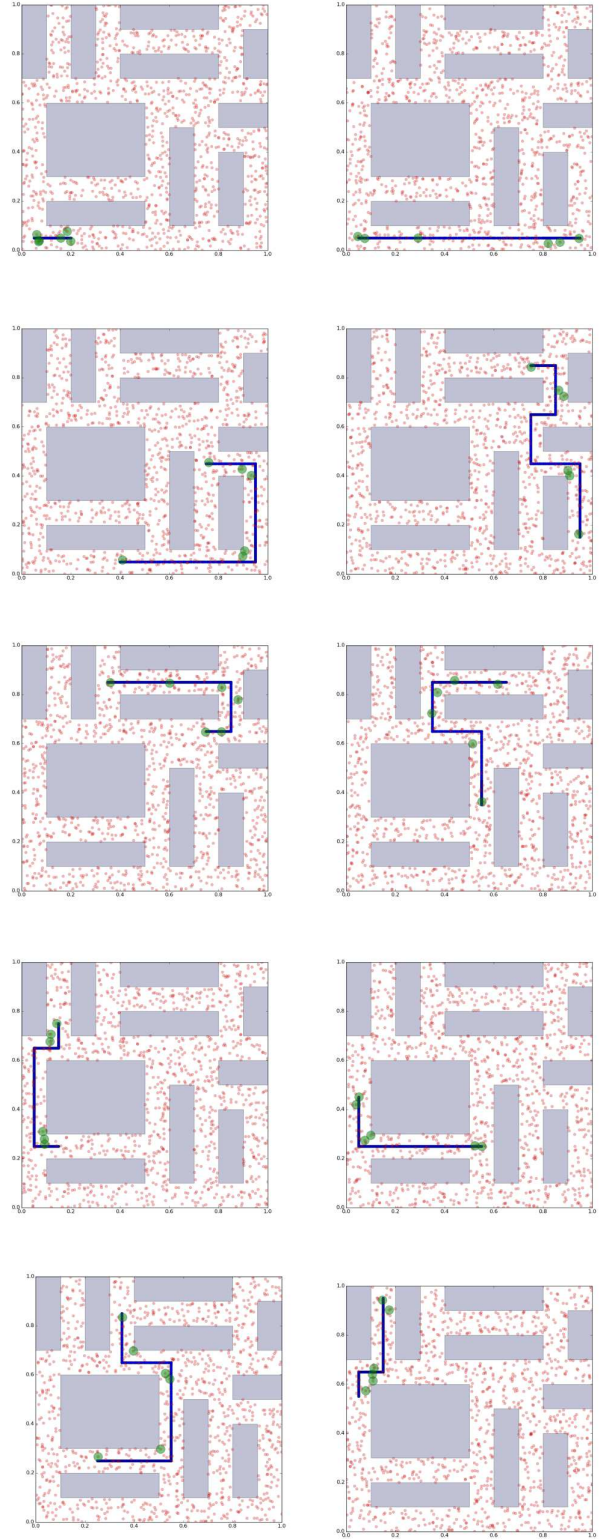}
\caption{Simulation study for the case of $12$ robots monitoring a dynamic parade route. The parade is considered to be taking place within a city--like environment consisting of $10$ building blocks. Camera-occlusions are accounted for, while the field of view of the camera that equips every robot is considered to be $175\textrm{deg}$. For this study $2048$ possible guard locations are sampled within the obstacle--free subset of the world. }
\label{fig:res1b}
\end{figure}

Similarly, Figure~\ref{fig:res1b} presents the results of the identical set--up with the exception of sampling $2048$ possible guard locations. As shown the results of the robots positioning are very similar for almost all iterations which indicates that as long as a \emph{sufficient} number of guard positions is sampled, then further enlargement of this sampling space will not tend to lead to significantly better solutions. On the other hand, computational time increases a lot as shown in Figure~\ref{fig:res_time}, a fact that further highlights the need for a good prior tuning of the amount of guard positions to be sampled. As the sampling of possible guard positions is uniform however, tuning this value is in general only about having one good reference value for a given environment and then scaling with the surface of free space. 

%
%

\begin{figure}[htbp]
\centering
  \includegraphics[width=0.99\columnwidth]{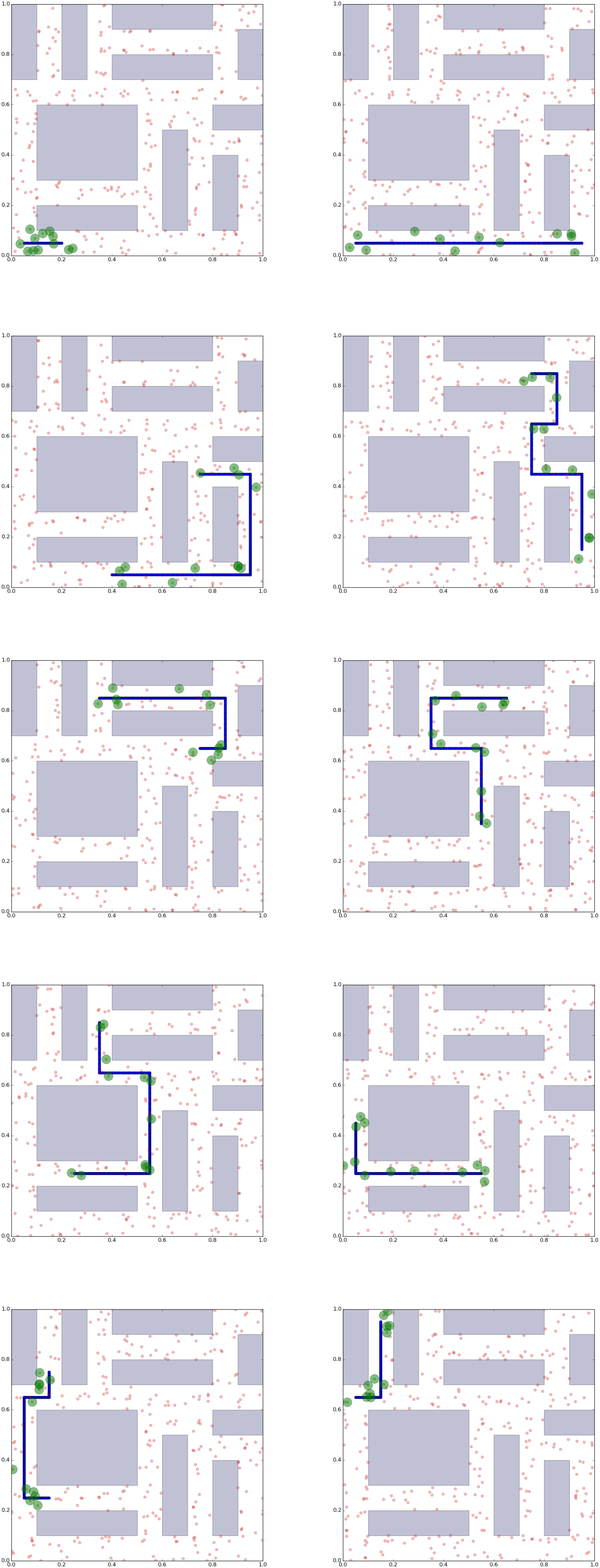}
\caption{Simulation study for the case of $12$ robots monitoring a dynamic parade route. The parade is considered to be taking place within a city--like environment consisting of $10$ building blocks. Camera-occlusions are accounted for, while the field of view of the camera that equips every robot is considered to be $175\textrm{deg}$. For this study $512$ possible guard locations are sampled within the obstacle--free subset of the world.}
\label{fig:res2}
\end{figure}

Figure~\ref{fig:res2} presents the same case but now with $12$ aerial robots. For this case, initially a total of $512$ possible guard positions are sampled within the obstacle--free subset of the workspace of the problem. As shown, the solution is characterized with more close pressence of robots around the parade route. Figure~\ref{fig:res_time} presents the computation characteristics of the solution per step of iteration.

%
%

\begin{figure}[htbp]
\centering
  \includegraphics[width=0.99\columnwidth]{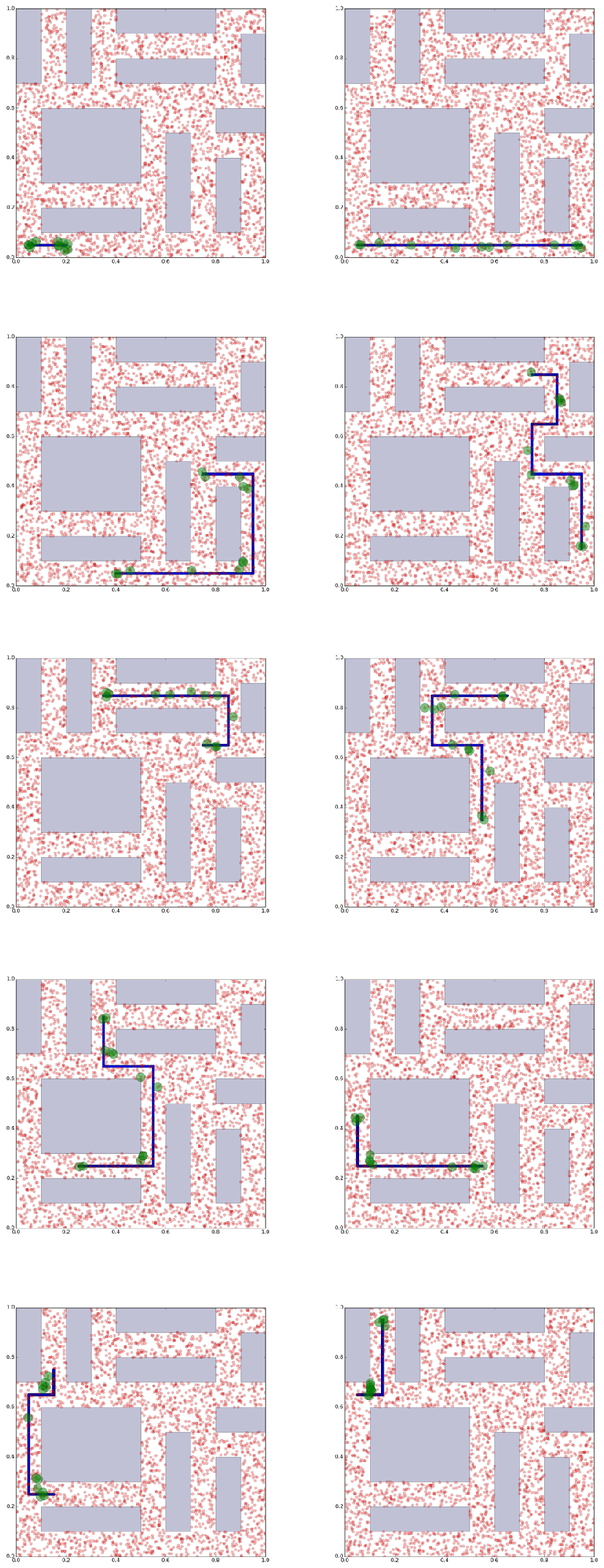}
\caption{Simulation study for the case of $12$ robots monitoring a dynamic parade route. The parade is considered to be taking place within a city--like environment consisting of $10$ building blocks. Camera-occlusions are accounted for, while the field of view of the camera that equips every robot is considered to be $175\textrm{deg}$. For this study $4096$ possible guard locations are sampled within the obstacle--free subset of the world. }
\label{fig:res2b}
\end{figure}

Similarly, Figure~\ref{fig:res2b} presents the results of the identical set--up with the exception of sampling $4096$ possible guard locations. Again the results of the robots positioning are similar for almost all iterations, which further denotes that very large sets of possible guard locations are not providing significant solution--quality benefits. On the other hand, computational time increases a lot as shown in Figure~\ref{fig:res_time}.

%
%

\begin{figure}[htbp]
\centering
  \includegraphics[width=0.90\columnwidth]{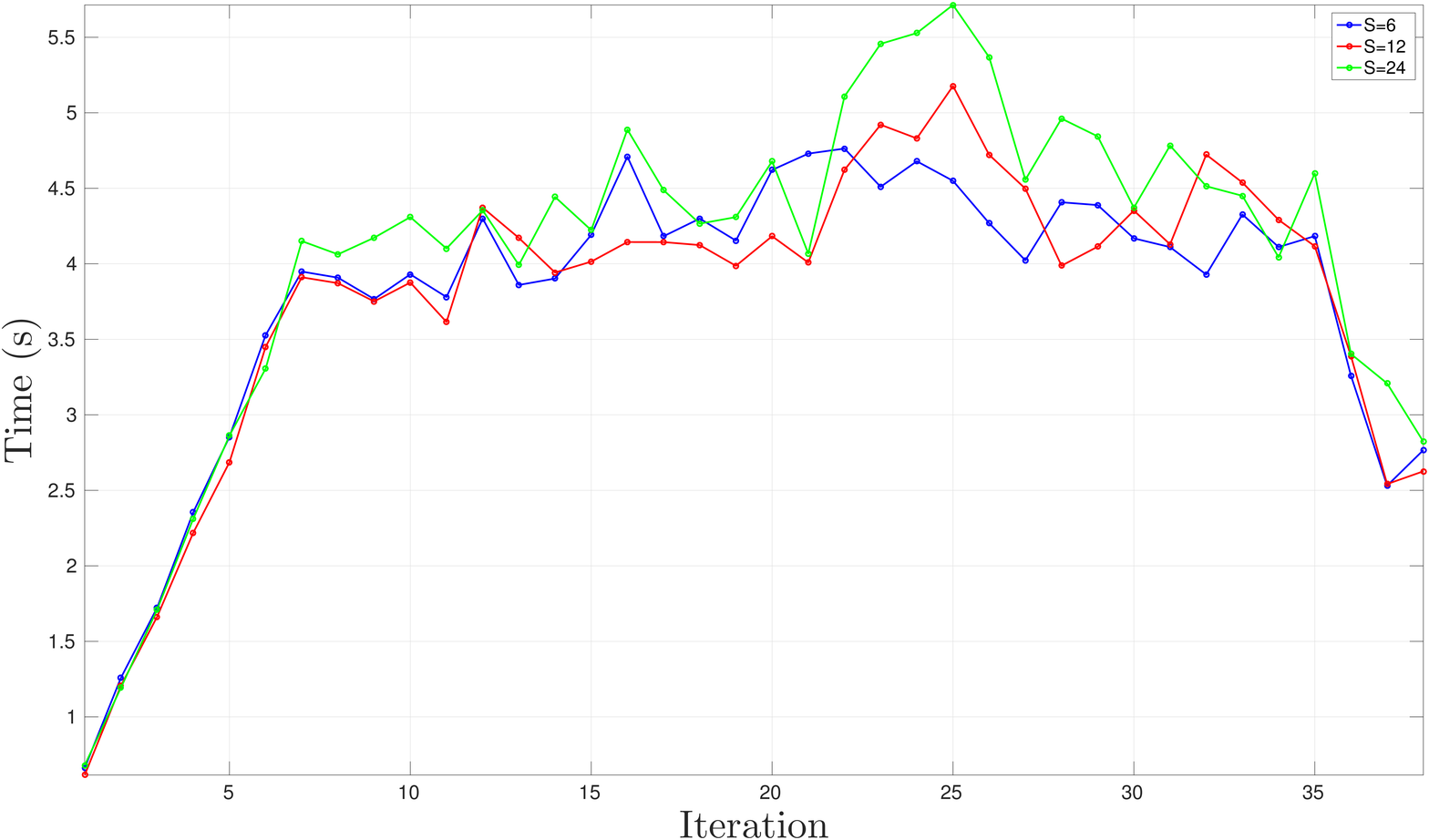}
\caption{Analysis of the computational cost per iteration of the algorithm for $6,12$ and $24$ robots given that the set of potential guard positions is set to the fixed value of $1024$. As shown the dynamics as well as the cost of the computation per iteration are similar regardless of the size of the team, a fact that highlights the scalability of the proposed approach.  }
\label{fig:timemultirob}
\end{figure}

Figure~\ref{fig:res_time} summarizes the computational properties of the algorithm for the above mentioned simulation cases. Furthermore, Figure~\ref{fig:timemultirob} presents the computational analysis of a set of studies with $6,12,24$ robots, while keeping the amount of potential sampled guard positions fixed to $1024$. As shown, the computational cost is very similar for the different robot teams both in the sense of the average value as well as of the evolution of it. This indicates the good scalability properties of the algorithm for arbitrary large teams of aerial robots. 

%
%

%
%

In summary, it was shown that the algorithm is able to deal with complex parade routes taking place in urban--like environments. Different sizes of robotic teams can be considered and the algorithm presents good computational scalability. Computation time is primarily affected by the size of the set of potential guard locations, which indicates that the size of the problem can influence the computation time. However, even in cases of very large potential guard location sets, the algorithm finds solutions within seconds - a performance considered to be sufficient given the large time scales of dynamic variations in social parades. At the current implementation of the algorithm, connection of subsequent optimal positions of the aerial robots team members relies on the nearest-neighbor concept as computed over collision--free trajectories. Future work will incorportate a full optimal solution employing Multiple--Vehicle--Routing--Problem solvers such as the implementation in~\cite{OPENVRP}.

\section{CONCLUSIONS}\label{sec:concl}

This technical report deals with the problem of positioning of a team of aerial robots such that they provide optimal coverage of a dynamically evolving parade taking place in an urban environment. The problem is solved iteratively over sampled representations of the parade route and it relies on convex approximates of the original noncovex problem. As the parade route is only as well covered as its least--covered point, the optimization objective is to place the aerial robots such that they maximize the minimum coverage over the points in the route at every time instant of it. Simulation studies verify the functionality of the algorithm, present its capacity to handle large robot teams and complex parade routes, as well as its low computational cost. 

\input{A_MSC2016.bbl}

\end{document}